# Reading Between the Lines: Combining Pause Dynamics and Semantic Coherence for Automated Assessment of Thought Disorder

Feng Chen[a,b,*], Weizhe Xu[a,b], Changye Li[a,b], Serguei Pakhomov[c], Alex Cohen[d,e], Simran Bhola[f], Sandy Yin[f], Sunny X Tang[f,g,h], Michael Mackinley[i], Lena Palaniyappan[j,k], Dror Ben-Zeev[b], Trevor Cohen[a,b]
[a]Department of Biomedical Informatics and Health Education, University of Washington, Box 358047, Seattle, WA 98195, USA
[b]Behavioral Research in Technology and Engineering (BRiTE) Center, Department of Psychiatry and Behavioral Sciences, 3751 W Stevens Wy NE, University of Washington, Seattle, WA 98195, USA
[c]College of Pharmacy, University of Minnesota, 308 SE Harvard St, Minneapolis, MN 55455, USA
[d]Department of Psychology, Louisiana State University, 103 Audubon Hall, Baton Rouge, LA 70803, USA
[e]Center for Computation and Technology, Louisiana State University, 1079 Digital Media Center, Baton Rouge, LA 70803, USA
[f] Northwell Health, 2000 Marcus Avenue, New Hyde Park, NY 11042, USA
[g]Feinstein Institutes for Medical Research, Institute of Behavioral Science, 350 Community Drive, Manhasset, NY 11030, USA
[h]Donald and Barbara Zucker School of Medicine, Department of Psychiatry, 500 Hofstra Blvd, Hempstead, NY 11549, USA
[i]London Health Sciences Centre, 339 Windermere Road, London, Ontario N6A 5A5, Canada
[j]Douglas Mental Health University Institute, Department of Psychiatry, McGill University, 6875 Boulevard LaSalle, Montreal, Quebec H4H 1R3, Canada
[k]Robarts Research Institute, Schulich School of Medicine and Dentistry, Western University, 1151 Richmond Street North, London, Ontario N6A 5B7, Canada
[*] Corresponding author
Email: fengc9@uw.edu, weizhexu@uw.edu, changyel@uw.edu, pakh0002@umn.edu, acohen@lsu.edu, sbhola2@northwell.edu, syin@northwell.edu, stang3@northwell.edu, michael.mackinley@lhsc.on.ca, lena.palaniyappan@mcgill.ca, dbenzeev@uw.edu, cohenta@uw.edu

**Abstract**

Formal thought disorder (FTD), a hallmark of schizophrenia spectrum disorders, manifests as incoherent speech and poses challenges for clinical assessment. Traditional clinical rating scales, though validated, are resource-intensive and lack scalability. Automated speech recognition (ASR) allows for objective quantification of linguistic and temporal features of speech, offering scalable alternatives. Furthermore, ASR-derived utterance timestamps provide access to pause dynamics, which are thought to reflect the cognitive processes underlying speech production. Yet, their added value beyond semantic measures remains insufficiently explored. In this study, we evaluated a scalable multimodal framework that integrates pause features with semantic coherence metrics across three datasets: naturalistic self-recorded diaries (AVH, n = 140 participants), structured picture descriptions (TOPSY, n = 72 participants), and dream narratives (PsyCL, n = 43 participants). Pause-related features were evaluated alongside established coherence measures using support vector regression (SVR) to predict clinical FTD scores. Models using pause features alone robustly predict manually rated FTD severity consistently across datasets. Integrating pause features with semantic coherence metrics enhanced predictive performance compared to coherence-only models, with late fusion yielding the most robust and consistent gains in all three datasets. On average across datasets, Spearman correlation increased from $\rho = 0.413$ for semantic-only models to $\rho = 0.455$ with late fusion. The performance gains from semantic and pause features integration held consistently across all contexts, though the nature of the most informative pause patterns was dataset-dependent. These findings suggest that both pause dynamics and semantic coherence reflect complementary aspects of thought disorganization. Our integration framework provides a scalable approach for refining the assessment of disorganized speech to advance automated speech analysis in psychosis.

**Key words:** Formal thought disorder; Natural language processing; Automated speech analysis; Schizophrenia spectrum disorders; Pause dynamics, Semantic coherence

**Highlights**

1. Pause features alone predict formal thought disorder across three datasets
2. Late fusion of pause and semantic features outperforms individual modalities
3. Task structure influences pause patterns in psychosis speech assessment
4. Automated multimodal analysis enhances clinical thought disorder detection
5. Cross-dataset validation demonstrates robust generalizability of approach

## 1. Introduction

Formal thought disorder (FTD) is a core clinical feature of schizophrenia spectrum disorders (SSD) and other psychotic conditions, characterized by disruptions in the organization and expression of thought (Roche et al., 2015; Sass & Parnas, 2017). This syndrome manifests through disorganized speech, encompassing phenomena such as derailment (oblique or unrecognizable connections between juxtaposed ideas), tangentiality (a disconnect from an initial question or topic), and semantic incoherence (weak or absent meaningful links between utterances). Such disorganized speech complicates clinical assessments, is associated with poorer social functioning, reduced treatment responsiveness, and worse long-term prognosis (Andreasen & Grove, 1986; Roche et al., 2015). Accurately identifying and quantifying these speech disruptions has traditionally relied on labor-intensive manual transcription and scoring using validated clinical scales, including the Thought and Language Disorder (TALD) scale, the Thought and Language Index (TLI), and the Assessment of Thought, Language, and Communication (TLC) scale (Andreasen, 1986; Kircher et al., 2014; Liddle et al., 2002). While these tools have been validated, their reliance on subjective interpretation, extensive training for assessors, and labor-intensive rating processes limits their reproducibility and scalability.

Early meta-analytic work has investigated speech-derived markers as objective indicators of psychopathology in schizophrenia spectrum disorders, with particular attention to temporal characteristics of speech production such as pauses, speech rate, and prosody (Cohen et al., 2014). In that meta-analysis, the authors identified pause-related measures as among the most robust acoustic correlates distinguishing individuals with schizophrenia from healthy controls. Consistent with these results, a recent systematic review and Bayesian meta-analysis reported that pause duration exhibits one of the largest average effect sizes in distinguishing individuals with schizophrenia from healthy controls, outperforming many other acoustic features (Parola et al., 2020). Both meta-reviews highlight substantial heterogeneity in effect sizes across studies, attributable to differences in speech tasks, recording conditions, and analytic choices. Importantly, while these prior studies primarily examined diagnostic group differences, they did not directly investigate the relationship between pause dynamics and clinician-rated formal thought disorder severity. This distinction is important, as case–control differences in speech timing do not necessarily imply sensitivity to severity in thought disorganization. Together, these findings suggest that while pause dynamics capture clinically meaningful disruptions in speech planning and execution, their predictive utility is strongly context-dependent and may vary across ecological settings and cognitive demands. Consequently, understanding how pause-related features behave across different speech elicitation paradigms remains a critical challenge for the development of reliable, generalizable speech-based markers.

Recent advances in automated speech analysis offer promising alternatives to these manual assessment methods by leveraging computational methods to quantify key characteristics of disorganized speech, such as linguistic coherence and temporal speech dynamics (Bedi et al., 2015; Elvevåg et al., 2007; Voleti et al., 2019). The advent of high-accuracy automatic speech recognition (ASR) systems, such as OpenAI's Whisper (Radford et al., 2023), has enabled the automated extraction of both textual content and precise temporal speech features, including start and end times for each segment of speech (Radford et al., 2023). Unlike manual transcription, ASR can rapidly process large volumes of audio recordings to generate time-aligned transcripts, thereby capturing not only the words spoken but also the temporal structure of speech. These timestamps enable the extraction of pause times—the silent intervals between utterances—which have long been thought to reveal important cognitive and communicative processes (Angelopoulou et al., 2024; Krivokapić et al., 2020; Matzinger et al., 2023). Specifically, irregular or prolonged pauses may reflect disruptions in the flow of thought (Çokal et al., 2019; Stanislawski et al., 2021), thereby offering an additional dimension for assessing speech organization in individuals with formal thought disorder.

Despite emerging interest and potential in using automated speech analysis (ASR) to facilitate automated assessment in psychiatry(Corcoran & Cecchi, 2020), most computational approaches to thought disorder have focused primarily on semantic coherence metrics, while temporal speech dynamics have received comparatively less attention (Bedi et al., 2015; Corcoran et al., 2020; Elvevåg et al., 2007; Xu et al., 2021; Xu et al., 2022). Semantic coherence metrics typically quantify the semantic relatedness between adjacent units of speech, thereby providing estimates of local coherence. Alternatively, global coherence measures assess how individual speech units relate to a summary representation of the broader context. For instance, cumulative centroid-based methods aggregate semantic embeddings from all preceding speech units to evaluate how well each subsequent unit aligns with the evolving context of the discourse (Xu et al., 2021). However, language-based coherence metrics alone may not fully capture the complexities of thought disorder, particularly because ASR errors can distort transcripts and spuriously depress semantic-similarity scores (Ciampelli et al., 2023). Additionally, because most sentence-embedding models are trained on general-domain text, they often miss the medical terms, local sayings, and neologisms that may show up in patients' speech (Berisha & Liss, 2024; El Boukkouri et al., 2019; Hitczenko et al., 2021). Furthermore, natural speech is typically not organized into discrete sentences, but ASR systems impose this artificial structure during transcription, which can introduce additional artifacts that may confound similarity measurements (Huang et al., 2023). Consistent with recent cross-linguistic and multi-dataset evaluations, the performance of semantic coherence measures has been shown to vary substantially depending on the choice of embedding model, aggregation strategy, and speech task, raising concerns about generalizability across clinical contexts (Parola, Lin, et al., 2023). Together, these factors may obscure the relationship between semantic similarity metrics and clinically meaningful disorganization, particularly when evaluated in isolation.

Pause features may provide complementary information to semantic coherence analyses. Research in non-clinical populations indicates that pauses often occur before semantically complex segments or when speakers shift topics both of which may require increased cognitive and linguistic planning (Krivokapić et al., 2020; Matzinger et al., 2023). In clinical populations, pause patterns have shown to be more variable, potentially reflecting an underlying challenge with organizing and producing fluent speech (Angelopoulou et al., 2024; Cohen et al., 2016; De Boer et al., 2023; Thakore et al., 2010). Recent studies have demonstrated that relatively simple prosodic characteristics—such as mean pause duration, number of pauses, and proportion of total speaking time—can significantly improve predictions of clinical severity in schizophrenia (De Boer et al., 2023; Thakore et al., 2010). Also, recent studies focusing on early psychosis have found that individuals may exhibit longer pause durations compared to controls, and that increased pause duration can be associated with greater severity of thought disorder (Dalal et al., 2025). Integrating pause features with coherence metrics may therefore enable models to better capture both timing and content of speech, potentially improving the detection of varying levels of disorganization and distinguishing task-specific speech dynamics, an important consideration given the diversity of clinical interviews, self-recorded narratives, and structured tasks used in psychiatric assessment (Cohen et al., 2016; Low et al., 2020).

Another vital consideration is the context-dependent nature of speech. For instance, individuals experiencing auditory verbal hallucinations (AVH) may exhibit distinct patterns of disorganized speech compared to participants who are not actively hallucinating. Moreover, speech characteristics can shift dramatically depending on the elicitation context—features that robustly capture disorganization in structured picture description tasks may underperform in open-ended diary entries, and controlled laboratory settings may yield different patterns than naturalistic environments. This systematic variation across contexts highlights the importance of a multi-dataset approach, which enables comparative evaluation of how pause summary statistics and semantic coherence metrics generalize across varying speech elicitation methods and clinical conditions.

Recent work has increasingly emphasized the potential benefits of multimodal speech analysis for characterizing schizophrenia-related language disturbances, while simultaneously highlighting challenges related to generalizability and overfitting. Studies combining linguistic and acoustic features have shown that multimodal models can outperform unimodal approaches in predicting or classifying psychosis-related outcomes, suggesting that temporal and semantic cues capture partially non-overlapping information about speech organization (Hansen et al., 2023). At the same time, large-scale evaluations have demonstrated that the performance of speech-based models is highly sensitive to analytic choices, including feature construction, aggregation strategies, and validation design, with models often failing to generalize across datasets, languages, or tasks (Parola, Lin, et al., 2023; Parola, Rybner, et al., 2023). These studies caution that apparent performance gains may reflect dataset-specific artifacts or overfitting rather than clinically robust markers, particularly in small samples or high-dimensional feature spaces. Together, this literature underscores the need for multimodal approaches that are evaluated across diverse speech contexts using carefully designed validation procedures, with explicit attention to how different modeling and fusion strategies influence robustness and interpretability.

In this study, we examine how pause features and semantic coherence metrics jointly contribute to automated assessment of FTD across three distinct speech datasets that involve people with SSD: (1) naturalistic self-recorded open-ended audio diaries from individuals with auditory verbal hallucinations (the AVH dataset) (Ben-Zeev et al., 2020), (2) structured picture-description interviews from people with first-episode psychosis (FEP) and healthy controls (the TOPSY dataset) (Van Dyken et al., 2024), and (3) dream narratives interview from participants with SSD (the PsyCL dataset) (Tang et al., 2023). These tasks represent different points along the spectrum of ecological validity and cognitive demand. This systematic variation enables us to examine how speech disorganization manifests differently across contexts and whether temporal features like pause dynamics show consistent or task-dependent relationships with clinical symptoms. We generated time-aligned transcripts capturing both linguistic content and pause intervals, leveraging WhisperX model for ASR (Bain et al., 2023). We then examined multiple predictive strategies, including unimodal models, early fusion via concatenation of features, and late fusion via post-hoc averaging of model results, to determine how best to merge temporal (pause) and textual (coherence) features.

We test three primary hypotheses: (i) pause features derived from ASR timestamps are independently predictive to clinical FTD scores; (ii) integrating pause features and coherence metrics yields more robust predictions than semantic features alone; (iii) the relative utility of multimodal fusion strategies depends on how temporal and semantic information are combined. By analyzing three datasets that differ in clinical stage, task structure and ecological validity, this work further explores the relationships between tasks used to elicit speech, and measures used to quantify coherence. Ultimately, through these efforts we aim to develop scalable, multimodal tools for objective speech analysis in psychosis.

The key contributions of this work are as follows: a) We systematically evaluate the utility of ASR-derived pause features for predicting clinical FTD scores across three distinct datasets, demonstrating their independent predictive power. b) We show that integrating pause dynamics with semantic coherence metrics consistently enhances predictive performance. This finding is particularly compelling given that the clinical ratings were often derived from text transcripts alone, meaning annotators had no access to temporal audio information. The performance gain thus demonstrates that pause features capture unique, non-semantic signals of thought disorganization that are complementary to semantic analysis. c) We compare multiple multimodal integration strategies and find that late fusion via model-level aggregation consistently outperforms early feature-level fusion across datasets, suggesting that temporal and semantic speech features provide partially independent information that is best combined at the decision level. d) We analyze how pause and coherence features vary by context, providing evidence for task- and illness-stage-dependent speech patterns in psychosis.

## 2. Method

This study follows a modular analysis pipeline that combines automatic speech recognition, extraction of semantic coherence and pause-based features, unimodal modeling, and multimodal integration to predict formal thought disorder severity. Figure 1 provides an overview of the full workflow, from audio preprocessing and feature extraction to regression modeling and evaluation.

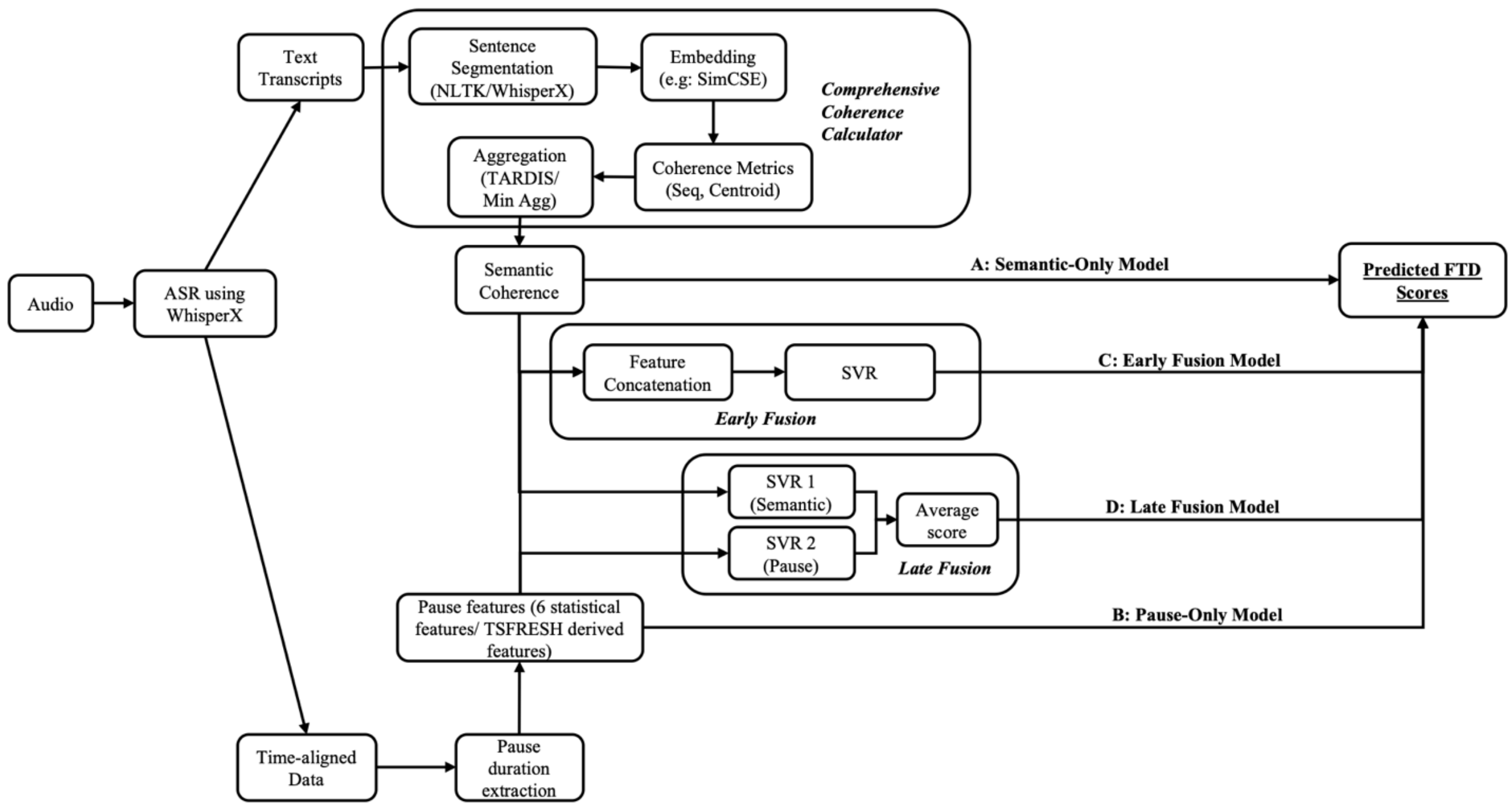

**Figure 1. Workflow diagram of the coherence prediction using semantic and pause features**. Pipeline shows feature extraction from WhisperX transcripts, calculation of pause and semantic metrics, and two integration strategies. Early fusion concatenates features before training a single model; late fusion averages predictions from separate pause and semantic models. ASR = Automatic Speech Recognition; SVR = Support Vector Regression; NLTK = Natural Language Toolkit; TARDIS = Time-series Augmented Representations for Detection of Incoherent Speech. FTD = Formal Thought Disorder.

*2.1. Dataset*

We utilized three datasets in this study, each containing recorded speech from participants from clinical populations experiencing psychosis-related symptoms. The characteristics of each dataset can be found in Table 1. The first dataset we analyzed was a smartphone "audio-diary" corpus that was originally gathered to study everyday experiences of AVH under the Mobile-RDoC project, in which adults who hear voices recorded short, free-speech "diaries" describing their voices and related events for 30 days using a bespoke app (Ben-Zeev et al., 2020). The full cohort comprised 384 participants recruited from 41 U.S. states. Over the course of the study, they submitted 4,809 recordings. For the present experiments, analyses were restricted to the subset of transcripts that had manual clinical annotations of derailment of speech. We used an annotated set of 310 recordings from 142 participants, developed by Xu et al. (Xu et al., 2021; Xu et al., 2022), comprising up to three randomly selected recordings of 30 seconds or more per participant, using data collected before October 18, 2019. Of these recordings, seven automated speech-to-text transcriptions containing no interpretable content were excluded, yielding a final dataset of 303 transcripts from 140 participants. Each transcript was manually annotated for derailment using the TALD scale. TALD scores ranged from 0 to 4, with 0 indicating no derailment, 1–2 indicating mild to moderate derailment, 3 indicating severe derailment, and 4 indicating incoherence. Each transcript was independently scored by two annotators. Any disagreements ≥ 2 points were re-rated, and the final score was calculated as the average of their ratings. The inter-rater agreement by quadratically weighted Kappa score was 0.71.

The second and third datasets were both obtained from PsychosisBank, a global initiative organized by the Discourse in Psychosis project to study thought, language, and communication disturbances in psychosis (Tan et al., 2023). The second dataset, referred to as the TOPSY dataset, consists of interview recordings from 106 participants who completed three picture description tasks. As the other two datasets include data from symptomatic individuals only, we focused our analysis on data from the 72 patients with FEP who had interview recordings with manual annotations of FTD scores. Thought disorganization was scored using the TLI, a measure that quantifies linguistic disturbances associated with psychosis. Since semantic coherence would not be expected across descriptions of different pictures, we segmented each participant's transcripts into three independent segments. Coherence and pause features were computed separately for each segment. This approach ensured that coherence analysis was conducted on transcripts related to a single topic.

The third dataset was taken from the PsyCL study. The data used here includes recordings from 43 participants with SSD who described their dreams. Each participant contributed one transcript, manually scored for thought disorganization using the TLC scale. This dataset provides focused context into thought disorganization within the context of dream descriptions. Of note, dream description represents a cognitively demanding task that requires participants to retrieve episodic memories of inherently bizarre and often illogical content, then organize this material into coherent speech (Nir & Tononi, 2010). While dream description may introduce task-specific pause patterns related to memory retrieval that may be independent of thought disorder, this cognitively demanding context provides a crucial contrast to the more naturalistic (diary) and structured (picture description) tasks, enabling us to differentiate between pause patterns reflecting underlying thought disorganization versus those attributable to specific cognitive demands.

These three datasets not only provide diverse speech contexts: encompassing naturalistic speech, structured interview responses, and descriptive narratives, but also represent different cohorts in terms of illness stage, with the TOPSY dataset specifically focusing on FEP, while the AVH and PsyCL datasets include individuals with broader schizophrenia spectrum disorders. Their inclusion allowed for a comprehensive investigation of thought disorganization across varying contexts, clinical conditions and potential illness stage.

**Table 1. Dataset Characteristics.** TALD = Thought and Language Disorder scale; TLI = Thought and Language Index; TLC = Thought, Language, and Communication scale. Threshold values represent clinical cutoffs for severe disorganization cases.

| **Dataset name** | **AVH** | **TOPSY** | **PsyCL** |
|---|---|---|---|
| **Task** | Audio-diary | Picture description | Dream recall |
| **Population** | Participants with auditory verbal hallucinations | Participants with first-episode psychosis | Participants with schizophrenia spectrum disorders |
| **Number of participants** | 140 | 72 | 43 |
| **Number of transcripts** | 303 | 216 | 43 |
| **FTD measurements** | TALD (0-4) | TLI (0-1) | TLC (0-4) |
| **Measured on** | Transcripts | Transcripts | Whole Interview Assessment |
| **Settings** | Monologue in naturalistic environment | Conversations in controlled environment | Conversations in controlled environment |
| **Recording length median** | 55.9s | 229.9s | 58.6s |
| **Pause proportion median** | 17.1% | 18.5% | 17.2% |
| **Pause counts median** | 10.0 | 40.0 | 9.0 |

*2.2. Automatic speech recognition and speaker diarization*

Automatic transcription across all datasets was performed using WhisperX, an open-source extension of OpenAI's Whisper automatic speech recognition (ASR) model (Bain et al., 2023; Radford et al., 2023). Whisper is a sequence-to-sequence Transformer-based ASR system trained on large-scale, weakly supervised multilingual speech data. Specifically, we used the whisper-large-v3 checkpoint, which is an updated pretrained model with improved transcription robustness and reduced hallucinations relative to earlier Whisper releases, that retains the same underlying architecture. WhisperX builds upon the Whisper model by adding forced alignment and speaker diarization components, enabling more precise segment-level and word-level timestamp estimation. In particular, WhisperX refines Whisper's segment boundaries through alignment-based post-processing and integrates speaker diarization via the Pyannote toolkit (Bredin, 2023), allowing for reliable separation of participant and interviewer speech in multi-speaker recordings.

For the TOPSY and PsyCL datasets, which involve interviewer–participant interactions, diarization was used to identify and retain only participant speech segments. Clinician speech was automatically excluded based on speaker labels and manually verified to ensure accuracy. For the AVH dataset, which consists of monologic self-recorded diaries, the same WhisperX pipeline was applied for consistency, although diarization was not required to separate speakers. The ASR pipeline proceeded as follows: (i) raw audio files were transcribed using WhisperX; (ii) time-aligned segments and word-level timestamps were generated; (iii) speaker diarization was applied where applicable; and (iv) participant-only speech segments were retained to construct time-aligned transcripts. These timestamps were subsequently used for both pause extraction and sentence segmentation in downstream feature extraction.
To assess transcription accuracy, we evaluated Word Error Rate (WER) and Character Error Rate (CER) on the AVH dataset, for which manually transcribed recordings with clinical annotations were available. The manually produced TALD transcripts served as reference text, and WhisperX-generated transcripts were treated as hypotheses. Prior to evaluation, transcripts were normalized by lowercasing and removing punctuation to ensure comparability. WER and CER were computed at the transcript level and then averaged across recordings; formal definitions are provided in Supplementary Note S1. We additionally compared WhisperX with the base Whisper model using the same evaluation protocol. WhisperX produced lower error rates, with WER decreasing from 21.3% (Whisper) to 14.5% (WhisperX) and CER decreasing from 17.0% to 9.2%. These values are reported to document transcription reliability and are not used as performance metrics in the predictive modeling analyses. Full details of the WER/CER computation are provided in the Supplementary Material.

*2.3. Aggregated semantic feature extraction*

**2.3.1 Overview and conceptual framework**

Semantic coherence features were originally developed to quantify coherence by measuring the degree to which successive units of speech are semantically related (Foltz et al., 1998), with lower coherence reflecting greater disruption in thought organization. Our approach follows prior computational work in psychosis, in which semantic coherence is estimated by computing similarity between vector representations of adjacent or context-related speech units derived from semantic vector representations derived using either Latent Semantic Analysis (LSA) or pretrained neural language models (Deerwester et al., 1990; Devlin et al., 2019; Mikolov et al., 2013). The underlying idea is that language that is loosely connected, reflecting disorganized thinking, should have lower semantic coherence scores between units of analysis. In practice, these scores are estimated by measuring the similarity between semantic vector representations in high-dimensional space, typically using the cosine metric (Bedi et al., 2015; Elvevåg et al., 2007; Just et al., 2019; Just et al., 2020; Tang et al., 2021).
In the current work, the semantic feature extraction pipeline comprises three stages: (1) segmentation of transcripts into speech units, defined as sentence-like units derived either from ASR-based punctuation (acoustically-derived WhisperX-split) or rule-based sentence splitting (grammatically-derived NLTK-split), which serve as the basic alignment units for pause-based analyses, as pauses are calculated between adjacent speech units in the processed transcripts; (2) computation of sentence-level semantic similarity scores using neural sentence embeddings; and (3) aggregation of these similarity scores into transcript-level features using either straightforward summary statistics or time-series feature extraction. All semantic features were computed using the Comprehensive Coherence Calculator (CCC), a modular framework developed for automated assessment of coherence in psychosis-related speech(Weizhe Xu, 2022).

**2.3.2. Sentence segmentation and embeddings**
Transcripts were segmented into sentence-like units using the sentence tokenizer from the Natural Language Toolkit (NLTK) (Bird, 2006; Bird et al., 2009). Because spontaneous speech is not naturally organized into grammatical sentences, this segmentation represents a computational approximation rather than a linguistic ground truth. To assess robustness to segmentation strategy, we additionally evaluated an alternative segmentation scheme in which boundaries between units were derived from WhisperX segment timestamps, which incorporate acoustic pause information. These units were then treated in the same way as the sentence-like units derived using NLTK.

Each grammatically- or acoustically-derived unit was embedded into a fixed-length semantic vector using pretrained sentence embedding models. In the main analyses, we report results using SimCSE (Simple Contrastive Learning of Sentence Embeddings), a contrastive-learning-based sentence embedding method that has been shown to consistently outperform standard BERT sentence representations on semantic similarity benchmarks (Gao et al., 2021). SimCSE embeddings were computed using a pretrained BERT-based backbone, without further task-specific fine-tuning for detection of disorganized thinking. A comparison with alternative embedding models implemented in CCC including BERT, Sentence-BERT is provided in the Supplementary Material.

**2.3.3. Sentence-level coherence computation**
Using sentence embeddings, CCC computes sentence-level semantic coherence scores using cosine similarity under three complementary formulations: (1) Sequential coherence (local coherence): cosine similarity between embeddings of consecutive sentences, capturing local topic continuity and abrupt topic shifts (Elvevåg et al., 2007). (2) Static centroid coherence (global coherence): cosine similarity between each sentence embedding and the centroid (vector average) of all sentence embeddings in the transcript, providing a fixed reference to overall discourse topic (Xu et al., 2021). (3) Cumulative centroid coherence (contextual coherence): cosine similarity between each sentence embedding and a centroid iteratively updated using all preceding sentences, modeling evolving discourse context and cumulative topic drift (Xu et al., 2021). Each formulation produces a sequence of sentence-level coherence scores for a given transcript.

**2.3.4. Transcript-level aggregation strategies**
To derive transcript-level semantic features from sentence-level coherence sequences, we employed two aggregation strategies, reflecting different modeling assumptions used in prior work.

*2.3.4.1. Minimum aggregation*
As a baseline approach, we computed the minimum coherence score within each transcript, yielding a single transcript-level value. This strategy reflects the assumption that the most incoherent segment dominates clinical judgments of disorganization and has been widely used in prior studies of automated coherence assessment (Bedi et al., 2015; Corcoran et al., 2018; Elvevåg et al., 2007). Because minimum aggregation produces a deterministic transcript-level score, it does not involve model fitting or cross-validation; performance is assessed via direct correlation with clinician ratings.

*2.3.4.2. Time-series feature extraction (TARDIS)*
As an alternative strategy, we adopted the Time-series Augmented Representations for Detection of Incoherent Speech (TARDIS) framework (Xu et al., 2022).Rather than collapsing coherence into a single summary statistic, TARDIS treats the sequence of coherence scores as a time series and applies automated feature extraction to capture temporal patterns across the entire discourse. Specifically, we used the TSFRESH package (Christ et al., 2018) to extract 764 statistical descriptors from each coherence time series, including measures of distributional shape, temporal structure, and frequency-domain properties. To avoid confounding coherence with transcript length, we excluded 25 length-dependent features, consistent with prior implementations of TARDIS (Xu et al., 2022). The resulting feature vectors represent transcript-level semantic coherence dynamics.

### *2.4. Pause feature extraction*

#### **2.4.1. Pause duration estimation**

Pause durations were estimated using time-aligned speech segments produced by WhisperX. Only pauses occurring between spoken segments were considered; leading and trailing silences at the beginning and end of recordings were excluded. WhisperX timestamp accuracy has been validated against manually verified word-level alignments with precision ranging from 84.1-93.2% using a 200-millisecond tolerance collar, providing confidence in our pause duration measurements (Bain et al., 2023). This process generated a list of pause durations for each transcript, which formed the basis for pause feature extraction.

#### **2.4.2. Aggregated pause features**

The first feature set comprised six *summary statistics* that succinctly summarized the pause patterns within the speech recordings: maximum, mean, median, and minimum pause durations, the total number of pauses, and the proportion of pause time relative to the total recording duration. These features provide a compact and clinically interpretable representation of temporal speech organization, enabling direct comparison with prior work on pause-related markers of thought disorder. The second feature set captured higher-order temporal structure in pause sequences using the same automated time-series feature extraction framework described in Section 2.3 to extract same 764 *temporal features* from TSFRESH package. A complete summary of all extracted clinical ratings, semantic coherence metrics, and pause-based temporal features, including scale, unit, and level of aggregation, is provided in Supplementary Table S1.

### *2.5. Modeling strategies*

To evaluate how temporal pause dynamics and semantic coherence jointly contribute to the prediction of FTD, we adopted a modeling framework that explicitly compares unimodal, feature-level (early fusion), and model-level (late fusion) approaches, as illustrated in Figure 1. This design allows us to assess (i) the independent predictive value of each modality, (ii) whether combining temporal and semantic information improves robustness, and (iii) how different integration strategies influence performance and generalizability across datasets. In addition, we compared multiple regression models to contextualize performance and reduce reliance on a single modeling assumption.

#### **2.5.1. Unimodal models**

As baselines, we trained unimodal models using (i) semantic coherence features only and (ii) pause-related features only. Semantic features included both minimum-aggregated coherence scores and time-series-derived features extracted using the TARDIS framework. Pause features included summary statistics and TSFRESH-derived temporal descriptors.

#### **2.5.2. Early fusion**

In the early fusion approach, pause features and semantic coherence features were concatenated into a single feature vector for each transcript. The combined feature vector was then used as input to a single regression model to predict annotator-assigned derailment scores.

#### **2.5.3. Late fusion**

In the late fusion approach, separate regression models were trained independently on pause features and semantic coherence features first. The final prediction for each transcript was obtained by averaging the predictions of the two unimodal models. This strategy preserves the distinct structure of each modality and reduces the risk that one feature set dominates the learning process due to scale or dimensionality differences, which has been shown to improve robustness in previous multimodal speech analysis efforts (Hansen et al., 2023).

#### **2.5.4. Model selection and training procedure**

Our primary regression model was Support Vector Regression (SVR) with a radial basis function (RBF) kernel, selected based on its prior use in automated assessment of thought disorder and its ability to model non-linear relationships in high-dimensional spaces (Xu et al., 2022). We additionally evaluated Ridge regression and partial least squares (PLS) regression, which provide linear baselines with different bias–variance tradeoffs and dimensionality-handling properties. We formulate the prediction task as regression rather than classification because the target clinical ratings are ordinal severity scales.

All regression models were implemented using scikit-learn in Python version 3.9.6. All models were trained and evaluated using leave-one-subject-out cross-validation (LOSO-CV) to prevent participant-level data leakage. In each fold, all transcripts from one participant were held out for testing, and models were trained exclusively on data from the remaining participants. Predictions were generated at the transcript level.

**2.5.5. Feature dimensionality control**

To mitigate potential overfitting arising from high-dimensional TSFRESH features, we conducted feature-selection sensitivity analyses within the training data of each LOSO fold using univariate F-statistics (f_regression). Results reported in the main manuscript are based on models using the top 200 selected features, which provided a favorable balance between predictive performance and robustness across datasets. Sensitivity analyses evaluating model performance across different feature dimensionalities are shown in Supplementary Figure S1.

*2.6. Evaluation*

The datasets used in this study incorporated different scoring systems for measuring thought and language disorganization. The raters contributing to the AVH dataset employed the TALD scale, those for the TOPSY dataset utilized the TLI, and those for the PsyCL dataset used the TLC scale. To account for these differences in scale structure and range, all models were trained and evaluated separately within each dataset.

**2.6.1. Evaluation metrics**

Performance was evaluated at the transcript level for all three datasets. We evaluated model performance using three complementary metrics. First, Spearman's rank correlation coefficient (ρ) was used as the primary evaluation metric to quantify the monotonic association between predicted scores and clinician ratings without assuming linearity or comparable scale intervals across instruments. Second, we report mean absolute error (MAE) to quantify absolute prediction error in the original rating space, providing an interpretable measure of predictive accuracy. Third, to evaluate model performance in identifying highly disorganized speech under class imbalance, we report the area under the precision–recall curve (AUPRC). AUPRC was chosen over ROC AUC because it is more informative when the positive (severe) class is relatively rare, as is typical in clinical speech datasets.

**2.6.2. Definition of severe disorganization**

Because TALD, TLI, and TLC differ in scale range and structure, and because no universally accepted clinical cutoffs exist across instruments, we adopted a distribution-based definition of severe disorganization. Within each dataset, transcripts (or participants) whose clinician ratings fell within the top 20% of the score distribution were labeled as "severe," and the remaining 80% as "non-severe." This approach provides a consistent severity prevalence across datasets and reduces arbitrariness introduced by scale-specific thresholds. Sensitivity of AUPRC to alternative quantile thresholds (50%–95%) is reported in Supplementary Figure S2.

*2.7. Secondary analyses*

In addition to predictive modeling, we conducted two secondary analyses to characterize relationships between temporal pause dynamics, semantic coherence, and task structure. These analyses were designed to probe underlying associations in the extracted features and to contextualize cross-task variability, rather than to evaluate predictive performance.

**2.7.1. Pause-coherence associations**

To assess whether longer pauses tend to precede greater semantic discontinuities in speech, we examined the association between pause duration and sentence-to-sentence semantic coherence at the transcript level. For each transcript, we aligned inter-sentence pause durations with sequential coherence scores computed between adjacent sentence units. Spearman's rank correlation coefficient was then calculated within each transcript to quantify the monotonic relationship between pause duration and local semantic coherence. This analysis was performed separately for each dataset and serves to characterize the extent to which temporal hesitation co-occurs with semantic shifts in discourse.

**2.7.2. Task-dependent variability**

To quantify how semantic coherence patterns vary across speech elicitation contexts, we examined task-dependent variability in sentence-level coherence measures. Specifically, we computed the coefficient of variation (CV) (Abdi, 2010) in sentence-level coherence across three metric categories: (1) all sentence coherence, (2) sequential coherence (local transitions), and (3) cumulative centroid coherence (global context aggregation). This approach enabled standardized comparison of coherence variability across datasets with different task demands, recording lengths, and discourse structures.

### *2.8. Statistical methods*

All statistical analyses were conducted in Python version 3.9.6. Model performance was assessed using Spearman's rank correlation (ρ), mean absolute error (MAE), and area under the precision–recall curve (AUPRC), as described in Section 2.6. Statistical comparisons between models were performed using the Wilcoxon signed-rank test, applied to paired performance estimates obtained across cross-validation folds. This nonparametric test was selected due to the non-normal distribution of performance metrics and the paired nature of the comparisons.

## 3. Results

### *3.1. Baseline performance: semantic-only vs pause-only models*

We first compared the predictive performance of semantic coherence features and pause-derived features independently, establishing unimodal baselines against which multimodal integration strategies were evaluated. Semantic-only models were derived from transcript-level coherence measures computed using sentence embeddings, while pause-only models relied exclusively on temporal pause features extracted from ASR-derived speech segments. All regression models were evaluated using leave-one-subject-out cross-validation to ensure speaker-independent assessment and to prevent within-speaker information leakage.

#### 3.1.1. Best performing semantic-only models

We systematically evaluated semantic-only models across the three datasets using NLTK-split ASR transcripts, following the framework proposed by Xu et al. (Xu et al., 2022). Specifically, we compared eight different sentence embedding models combined with three coherence aggregation strategies—sequential, static centroid, and cumulative centroid—resulting in 24 semantic coherence variants per dataset. For each variant, transcript-level representations were constructed using the TARDIS framework and trained using SVR. Detailed correlations between each semantic coherence variant and the corresponding clinical ratings are reported in the Supplementary Material Table S2.

Across all three datasets, SimCSE emerged as the best-performing embedding model. Using SimCSE embeddings, the optimal aggregation strategy varied by dataset: cumulative centroid aggregation yielded the strongest correlations in the AVH (ρ = 0.544 vs. 0.513 sequential, 0.526 static) and PsyCL datasets (ρ = 0.349 vs. 0.261 sequential, 0.233 static), whereas static centroid aggregation performed best in the TOPSY dataset (ρ = 0.345 vs. 0.159 sequential, 0.275 cumulative). These best-performing SimCSE-based coherence metrics were therefore used for all subsequent semantic and multimodal analyses.

To contextualize these results relative to prior work, we compared the TARDIS-based SVR model with the widely used minimum aggregation baseline using the same best-performing SimCSE embedding (Figure 2). Across all three datasets, TARDIS-based model consistently outperformed minimum aggregation, indicating that modeling sentence-to-sentence semantic dynamics provides a more informative representation of formal thought disorder severity than reliance on a single extreme value. We additionally compared the performance of support vector regression (SVR) with two alternative regression models—ridge regression and partial least squares (PLS)—using the same best-performing semantic features (Figure 2). SVR consistently achieved the highest correlations across datasets, and was therefore retained as the primary regression model for subsequent analyses.

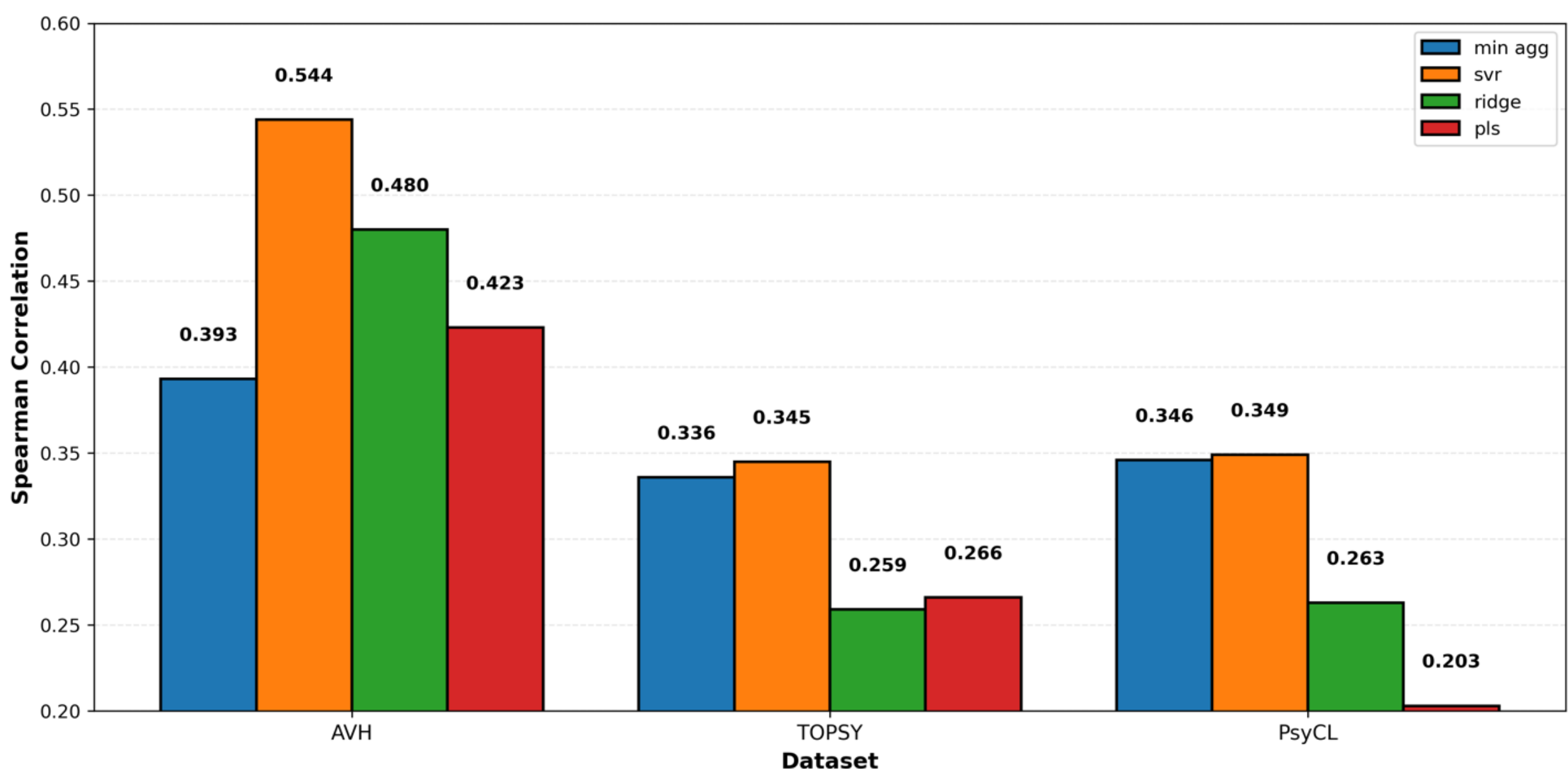

**Figure 2. Comparison of semantic-only prediction performance across aggregation and regression strategies.** Performance (Spearman correlation) is shown for minimum aggregation, SVR, ridge regression, and partial least squares (PLS) using the best-performing semantic coherence features for each dataset.

To assess the impact of transcript segmentation on semantic coherence estimation, we further compared NLTK-based sentence segmentation with WhisperX-derived utterance segmentation. As shown in Table 2, NLTK-based segmentation consistently outperformed WhisperX-based segmentation across all three datasets. This suggests that ASR-derived utterance boundaries did not provide additional benefit over punctuation-based sentence segmentation for semantic coherence modeling in this setting. Accordingly, NLTK-based segmentation was used as the semantic feature input for all fusion models.

**Table 2. Average performance comparison of unimodal, and multimodal models for FTD severity using SVR for three datasets.** All results are obtained using leave-one-subject-out cross-validation, with feature selection performed within each training fold (top 200 features). Performance is reported using Spearman's rank correlation coefficient (ρ; primary evaluation metric), mean absolute error (MAE – unlike the other metrics, lower values indicate better performance), and area under the precision–recall curve (AUPRC). †Cumulative centroid aggregation. ‡Static centroid aggregation. Bold values indicate the highest Spearman ρ within each dataset.

| | | **Average performance across three datasets** | | |
|---|---|---|---|---|
| | | Spearman Rho | MAE | AUPRC |
| Semantic-only models | NLTK-split | 0.413 | 0.538 | 0.512 |
| | WhisperX-split | 0.348 | 0.547 | 0.507 |
| Pause-only models | summary statistics | 0.367 | 0.551 | 0.528 |
| | TSFRESH-based | 0.315 | 0.568 | 0.511 |
| Fusion models | Early fusion | 0.418 | 0.534 | 0.528 |
| | Late fusion | **0.455** | **0.524** | **0.553** |

### 3.1.2. Pause-only models' performance is comparable to semantic-only models

Pause-only models showed consistent predictive value across datasets, with performance depending on the feature representation. Table 2 summarizes unimodal semantic-only and pause-only performance across datasets. In AVH, pause summary statistics achieved ρ = 0.498 (MAE = 0.558, AUPRC = 0.498), and the higher-dimensional pause representation (TSFRESH-based) yielded a similar correlation (ρ = 0.487, MAE = 0.561, AUPRC = 0.491). In TOPSY, pause summary statistics were weaker (ρ = 0.286, MAE = 0.298, AUPRC = 0.561), whereas TSFRESH-based pause features substantially improved performance (ρ = 0.402, MAE = 0.277, AUPRC = 0.621) and outperformed the semantic-only baseline in this dataset (Table 2). In PsyCL, pause summary statistics achieved ρ = 0.316 (MAE = 0.798, AUPRC = 0.524), while TSFRESH-based pause features performed poorly (ρ = 0.057, MAE = 0.865, AUPRC = 0.422), suggesting that the higher-dimensional pause representation does not generalize well to this task. Notably, the relative effectiveness of different pause feature representations varied by dataset, likely reflecting differences in speech task demands and illness stage. Based on these unimodal results, time summary statistics were selected for fusion models in the AVH and PsyCL datasets, while TSFRESH-based temporal features were retained for the TOPSY dataset.

Taken together, the semantic-only and pause-only baselines establish that both modalities independently encode meaningful signals related to thought disorder severity. These findings provide a clear and interpretable foundation for evaluating whether and how combining semantic coherence with pause dynamics improves prediction performance in subsequent multimodal analyses.

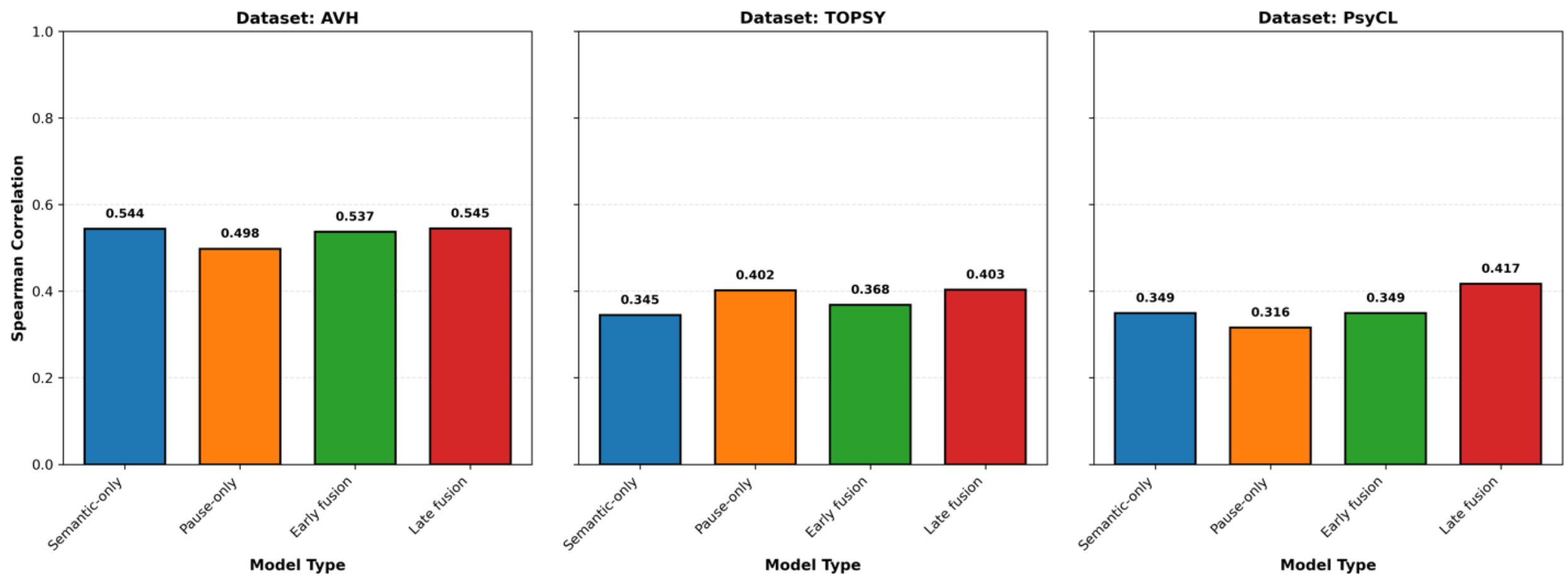

**Figure 3. Prediction performance (Spearman correlation) of unimodal and multimodal models across datasets.** Results are shown for AVH (left), TOPSY (center), and PsyCL (right). For multimodal models, early fusion represents feature-level concatenation, while late fusion represents model-level aggregation.

*3.2. Multimodal integration improves prediction of FTD severity*

We next evaluated whether combining semantic coherence with pause-derived features improves prediction of formal thought disorder severity relative to unimodal baselines. Across datasets, multimodal models achieved the strongest overall performance, with late fusion yielding the highest average Spearman correlation ($\rho = 0.455$) compared with semantic-only ($\rho = 0.413$) and pause-only models ($\rho = 0.367$). Figure 3 provides an overview of unimodal and multimodal model performance across datasets. This indicates that combining semantic coherence and pause dynamics provides a more robust signal of formal thought disorder severity than either modality alone. Comprehensive evaluation metrics are presented in Table 2. Based on the unimodal results reported in Section 3.1, we selected pause time summary statistics for the AVH and PsyCL datasets and TSFRESH-based pause representations for the TOPSY dataset. For each dataset, we evaluated two multimodal integration strategies: early fusion via feature-level concatenation and late fusion via model-level aggregation, using the same leave-one-subject-out evaluation protocol.

In the AVH dataset, integrating pause time statistics with semantic coherence resulted in modest but consistent improvements over unimodal models. Late fusion achieved the highest overall performance, with a Spearman correlation of $\rho = 0.545$ (MAE = 0.539, AUPRC = 0.487), slightly outperforming both the semantic-only baseline ($\rho = 0.544$) with more substantial improvements over the pause-only baseline ($\rho = 0.498$). Early fusion also improved upon the pause-only model ($\rho = 0.537$, MAE = 0.543), but did not exceed late fusion. These results indicate that pause dynamics and semantic coherence provide complementary information in the AVH diary narratives, with model-level integration yielding the most stable gains.

In the TOPSY dataset, late fusion model achieved the comparable performance ($\rho = 0.403$, MAE = 0.282, AUPRC = 0.602), improving considerably upon the semantic-only model ($\rho = 0.345$) and slightly exceeding the pause-only model ($\rho = 0.402$). Early fusion also outperformed the semantic-only baseline ($\rho = 0.368$, MAE = 0.285), but remained inferior to late fusion. The larger gains observed in TOPSY suggest that combining semantic coherence with richer temporal pause representations is particularly effective in structured picture-description tasks.

In the PsyCL dataset, multimodal integration again improved prediction when pause time statistics were combined with semantic coherence. Late fusion achieved a Spearman correlation of $\rho = 0.417$ (MAE = 0.752, AUPRC = 0.571), outperforming both the semantic-only ($\rho = 0.349$) and pause-only ($\rho = 0.316$) models. In contrast, early fusion did not yield additional gains beyond the semantic-only baseline ($\rho = 0.349$, MAE = 0.773), highlighting a consistent advantage of model-level over feature-level integration in this dataset.

Across all three datasets, multimodal models consistently matched or exceeded the best unimodal baselines, with late fusion outperforming early fusion in every case, as shown in Figure 3. Beyond the best-performing configurations reported here, fusion models improved Spearman correlation relative to semantic-only baselines across all 24 combinations of embedding models and coherence aggregation strategies evaluated, with consistent gains observed across datasets (see Supplementary Table S3), demonstrating that the advantage of late fusion is robust to embedding and aggregation choice. This pattern suggests that pause dynamics and semantic coherence capture partially independent aspects of speech organization related to thought disorder severity that are best integrated at the decision level rather than through direct feature concatenation. To assess robustness across tasks, we further evaluated cross-dataset generalization by training models on two datasets and testing on the held-out third dataset. As shown in Supplementary Table S4, late fusion still achieved the highest average Spearman correlation across cross-dataset splits compared with unimodal models.

*3.3. Task- and embedding- dependent associations between pause duration and semantic coherence*

Using WhisperX-derived segmentation, both sequential coherence and pause duration result in a single measurement for each pair of successive utterances, permitting an assessment of the correlation between measurements from each of these modalities. We therefore assessed the association between pause duration preceding each sentence and the semantic similarity between that sentence and the preceding one across datasets and embedding methods. As shown in Table 3, in the AVH and PsyCL datasets, slight negative correlations emerged between pause duration and sequential coherence measures. This indicated that *longer* pauses were associated with *reduced* sentence-to-sentence coherence, suggesting that longer pauses accompany semantic shifts in open-ended speech. Strikingly, the TOPSY dataset showed no significant correlation between pause duration and coherence measurements, suggesting that picture description tasks may alter variability in speech patterns and/or the flow of ideas, obscuring the link between the timing and semantics of speech. These weak correlations provide direct evidence that pause features and semantic coherence capture largely independent aspects of speech disorganization, supporting our finding that their integration consistently improves predictive performance through complementary rather than redundant information.

In addition, Table 3 also illustrates that the strength of pause-coherence associations depend on how semantic similarity is computed. Even within the same dataset, correlation magnitudes varied across embedding models, underscoring that sentence-level coherence estimates are sensitive to representational choices. This embedding dependence aligns with recent findings that different implementations of semantic coherence can yield divergent results and reinforces the need to interpret coherence-based measures in light of their computational formulation (Parola, Lin, et al., 2023).

**Table 3. Spearman correlations between pause durations and sequential semantic similarity at the sentence level across embedding methods.** Correlations computed between pause time preceding each sentence and semantic similarity to previous sentence. Bold values indicate statistically significant correlations ($p < 0.05$).

| | **AVH** | | **TOPSY** | | **PsyCL** | |
|---|---|---|---|---|---|---|
| Sentence-level semantic vector representation | ***Spearman Rho*** | ***p-value*** | ***Spearman Rho*** | ***p-value*** | ***Spearman Rho*** | ***p-value*** |
| Skip-gram embeddings | **-0.078** | **<0.01** | -0.054 | 0.23 | **-0.162** | **<0.01** |
| TF-IDF weighted skip-gram embeddings | -0.016 | <0.285 | 0.064 | 0.153 | **-0.084** | **<0.01** |
| Sum of the BERT token embeddings | **-0.061** | **<0.01** | 0.040 | 0.370 | -0.042 | 0.070 |
| Second-to-last layer of BERT hidden state embeddings | **-0.099** | **<0.01** | -0.002 | 0.957 | **-0.066** | **<0.01** |
| BERT CLS token embeddings | **-0.136** | **<0.01** | -0.050 | 0.269 | **-0.110** | **<0.01** |
| Sentence-BERT embeddings | **-0.107** | **<0.01** | 0.016 | 0.723 | **-0.154** | **<0.01** |
| SimCSE embeddings | **-0.084** | **<0.01** | -0.083 | 0.067 | **-0.110** | **<0.01** |
| DiffCSE embeddings | **-0.076** | **<0.01** | -0.031 | 0.482 | **-0.135** | **<0.01** |

To contextualize these findings, we analyzed the CV in sentence-level coherence across three categories of metrics based on SimCSE sentence embedding: sequential coherence, static centroid and cumulative centroid methods. As shown in Table 4, TOPSY exhibited the lowest variation across all coherence metric types, especially for sequential (28.79%) and static centroid (18.15%), consistent with the structured format of the picture description task, which limits topic divergence. Conversely, AVH and PsyCL, with open-ended prompts, demonstrated higher variability. With AVH, the CV for sequential coherence was 40.46%, while its cumulative centroid methods showed reduced variability (28.49%). With the PsyCL dataset we found intermediate CV values compared with other two datasets, as one might anticipate given that the task of dream description is more constrained than the audio diary task, but not as constrained as a picture description. With respect to methodological differences, cumulative centroid methods consistently yielded lower CV values compared to sequential approaches across all datasets. In AVH, cumulative centroid methods reduced variability by 11.97%, in PsyCL, by 13.84%, and in TOPSY, by 10.64%. This pattern suggests that cumulative centroid methods, which aggregate semantic context across sentences, stabilize coherence measurements by mitigating localized disruptions, which may explain in part their apparent advantages for the detection of thought disorganization.

**Table 4. Coefficient of variation (CV) for sentence-level semantic coherence across coherence calculation methods using SimCSE embeddings.** CV = standard deviation/mean × 100%. Lower CV indicates more consistent coherence within transcripts.

| | AVH | TOPSY | PsyCL |
|---|---|---|---|
| Sequential | 40.46% | 28.79% | 43.53% |
| Static Centroid | 41.38% | 31.14% | 39.01% |
| Cumulative Centroid | 28.49% | 18.15% | 29.69% |

These findings highlight the critical role of task structure in shaping the relationship between pause times and coherence measures. Structured tasks, such as TOPSY's picture description, suppress natural speech variability, reducing the sensitivity of coherence measures to disruptions. Conversely, unstructured tasks, such as audio diaries and dream descriptions, amplify variability and enhance the detection of coherence disruptions linked to thought disorganization. These patterns are strongly dataset-dependent, with data from each task exhibiting distinct relationships between pause times and coherence metrics. For clinical applications, this underscores the importance of aligning both task design and analytical frameworks to strengthen signal related to the construct of interest.

*3.4. Associations between pause time summary statistics and clinical ratings*

To explore the extent to which individual features are associated with clinical ratings, we further analyzed the Spearman correlations between the six pause summary statistics and FTD scores across the three datasets, revealing task-dependent patterns, as shown in Table 5. These findings are further contextualized by variability in sentence-level coherence (Table 4). In the AVH dataset, TALD scores correlate strongly with number of pauses ($\rho = 0.507$, $p < 0.05$) and transcript length ($\rho = 0.437$, $p < 0.05$), as participants with higher ratings for disorganization tended to have longer transcripts. With respect to other features, the TOPSY dataset (structured picture descriptions from FEP cohort) shows a paradoxical pattern: while the number of pauses correlates positively with TLI scores ($\rho = 0.524$, $p < 0.05$), mean pause duration ($\rho = -0.271$, $p < 0.05$) and pause proportion ($\rho = -0.200$, $p < 0.05$) exhibited inverse but weaker relationships. This likely reflects compensatory behavior in structured tasks—participants with disorganization, particularly in earlier stages of psychosis, may still pause more frequently, but shorten individual pauses to maintain coherence, a strategy supported by TOPSY's low variability in cumulative centroid coherence (CV = 14.99%), which stabilizes measurements of global coherence. In the PsyCL dataset, all pause features showed weak, non-significant correlations with TLC scores (e.g., number of pauses: $\rho = 0.126$, $p > 0.05$), perhaps due to the limited sample size ($n = 43$), but still in the same directions as AVH dataset.

These results indicate different relationships between task structure and how pauses relate to thought disorganization. Structured tasks like the picture description task used to elicit data for TOPSY may suppress variability in topic, reducing sensitivity to disorganization but revealing a tendency toward pressure of speech, a separate manifestation of thought disorder, through inverse pause duration correlations. Open-ended tasks like the audio diary task used to elicit the AVH transcripts provide opportunities for participants to change topics an increase in the variability of semantic coherence measures, linking pauses directly to disorganization, while tasks involving recollection such as dream descriptions may induce pauses related to memory retrieval as well. The complementary role of pause and coherence metrics is evident in their combined ability to disentangle these patterns: pauses index behavioral adaptations, while coherence metrics capture semantic disruptions. This dual perspective, particularly when using cumulative centroid methods to reduce variability, enhances alignment with human ratings across contexts. These findings further support that our methods of integrating pause dynamics and coherence measures generalize across various communication styles, which balance the ecological validity with clinical interpretability in disorganization assessment.

**Table 5. Spearman correlations between individual pause summary statistics and thought disorder severity scores.** Bold values indicate statistically significant correlations ($p < 0.05$).

| **Spearman Rho** | **Max pause time** | **Mean pause time** | **Median pause time** | **Min pause time** | **Number of pauses** | **Pause proportion** |
|---|---|---|---|---|---|---|
| AVH (TALD) | **0.135** | **-0.154** | **-0.185** | **-0.147** | **0.507** | 0.013 |
| TOPSY (TLI) | **-0.161** | **-0.271** | **-0.230** | -0.06 | **0.524** | **-0.200** |
| PsyCL (TLC) | 0.076 | 0.034 | 0.070 | 0.012 | 0.126 | 0.055 |

### *3.5. Using ASR instead of manual transcription*

We evaluated the performance of ASR using WhisperX package compared to manual transcription, focusing on the AVH dataset as a benchmark. The WER and CER were calculated to assess the accuracy and consistency of the ASR-generated transcripts. The WER was 14.5%, and the CER was 9.2%. These rates represent a notable improvement over those obtained using the base OpenAI Whisper model directly on the same data, which resulted in a WER of 21.3% and CER of 17.0%. These reduced error rates achieved by WhisperX, reflecting substantially improved transcription accuracy and fewer hallucinations compared to the base Whisper model, are considered crucial for the integrity of our downstream analyses focusing on semantic coherence and pause-time dynamics. While the impact of ASR errors on downstream clinical classification can be complex, with some research suggesting that certain error patterns might even contain useful diagnostic signals in tasks like dementia detection, (Li et al., 2024) our aim was to minimize the impact of transcription inaccuracies and to assess how intrinsic speech features relate to clinically-rated thought disorganization. For analyzing semantic coherence and precise pause timings, higher transcription fidelity is generally presumed to be beneficial. To further investigate the relationship between ASR performance and thought disorganization, we assessed the correlation between WER and the TALD scores. A statistically significant positive correlation was observed ($\rho = 0.24$, $p < 0.01$), suggesting that higher levels of thought disorganization are associated with increased transcription errors. This finding aligns with the expectation that disorganized speech patterns, such as derailment or incoherence, may pose greater challenges for ASR systems. These results indicate that Whisper-based ASR provides a robust alternative to manual transcription, with transcription accuracy that is adequate for the quantification of incoherence. However, the positive correlation between WER and FTD scores highlights the need for further refinement of ASR models to better handle speech patterns associated with severe thought disorganization. Future work could explore fine-tuning ASR models on datasets with higher levels of disorganization or incorporating domain-specific adaptations to improve transcription accuracy in clinical contexts. These findings support the feasibility of using ASR for large-scale studies of thought disorganization, while also underscoring the importance of considering the impact of speech characteristics on transcription quality.

## 4. Discussion

This study demonstrates that pause features significantly enhance the automated detection of thought disorganization when integrated with semantic coherence metrics. Our findings reveal that pause features independently predict clinical FTD scores and complement semantic coherence measures, offering a multimodal framework that improves alignment with human ratings across diverse speech contexts. These results clarify the features marking the clinical manifestation of disorganized speech in psychosis and provide actionable insights for refining clinical assessment tools, in line with the emerging agenda of employing computational linguistics in clinical psychiatry (Corona Hernández et al., 2023).

A key finding is that pause features including summary statistics (maximum, mean, median, minimum pause durations, pause frequency, and proportion of pause time), and TSFRESH-based high-dimensional features robustly predicted FTD scores across all three datasets. Of critical importance is the fact that the TALD (for AVH dataset) and TLI (for TOPSY dataset) ratings were derived from manual transcripts with raters having no access to pause information. This suggests that pauses are not merely artifacts of speech production but reflect intrinsic cognitive disruptions that correlate with manifestations of thought disorganization that are apparent to human observers when they examine transcribed speech outputs. The work from Lesh and Sharpe (Lesh et al., 2011; Sharpe et al., 2025). suggest that impaired context processing, which reflects the inability to maintain and utilize task-relevant information, underlies cognitive disruption in schizophrenia. Such deficits may manifest behaviorally as more pauses, as individuals struggle to integrate prior context with ongoing speech, resulting in fragmented output (Hart

& Lewine, 2017). This mechanistic overlap suggests that pause frequency captures disruptions in the dynamic interplay between cognitive control and language production, a hallmark of thought disorder.

Interestingly, in TOPSY, the inverse relationship between mean pause duration (ρ = -0.271) and proportion of pause time (ρ = -0.200) with TLI may indicate compensatory mechanisms: untreated patients with psychosis and more prominent disorganization might pause more frequently but shorten individual pauses to adhere to task demands, masking overt incoherence. Though this may appear counterintuitive, these findings are supported by Matthews et al. (Matthews et al., 2014), who demonstrated that, under conditions with minimal working memory (WM) maintenance demands, individuals with schizophrenia exhibit enhanced visuospatial imagery manipulation (evidenced by faster response times) despite an overall WM maintenance deficit.

The integration of pause features with semantic coherence metrics consistently improved alignment with human judgement compared to unimodal models using either feature type alone across all datasets, particularly through late fusion strategies which average two independent models' outputs. In the AVH dataset, the best-performing multimodal model achieved a Spearman correlation of ρ = 0.545, exceeding the strongest correlation reported for semantic coherence alone in prior work using the same dataset and evaluation framework, ρ = 0.465 in Xu et al. (Xu et al., 2022). At the same time, recent work has emphasized that the absolute magnitude of coherence–symptom correlations is highly sensitive to dataset characteristics, task structure, and the specific implementation of coherence measures(Parola, Lin, et al., 2023). Consistent with this observation, we find that absolute correlation values vary across datasets, whereas the direction and consistency of multimodal gains replicate across three independent datasets with distinct tasks and clinical populations. Taken together, these findings suggest that the primary advantage of multimodal integration lies not in large absolute performance increases, but in providing a robust framework that consistently matches or exceeds the best unimodal configuration without requiring a priori knowledge of which feature set will be most informative in a given task or dataset.

These results demonstrate that pause features and transcript semantic features capture complementary aspects of disorganization. This synergy is likely rooted in their distinct mechanistic origins: coherence metrics reflect semantic planning deficits, while pauses index disruptions in speech motor control or cognitive load. Averaging the result from two models preserves the interpretability of each feature set while leveraging their combined predictive power. Importantly, the performance improvements were not confined to a single embedding choice or coherence formulation. Across all 24 semantic configurations evaluated (8 embeddings × 3 coherence calculation approaches), integrating pause features increased Spearman correlation relative to semantic-only models in all three datasets (STable 1), with statistically significant improvements confirmed via Wilcoxon signed-rank tests. These consistent and statistically supported gains validate the added value of pause-time features when combined with semantic-only models. This robustness is particularly notable given the embedding-dependent variability observed in coherence estimates, suggesting that pause-based temporal features provide a stabilizing signal that is less sensitive to representational choices.

Furthermore, our results underscore that pause patterns and coherence metrics are highly task-dependent, and may also reflect differences related to the stage of psychosis among the participant cohorts. For instance, the TOPSY dataset predominantly included individuals with first-episode psychosis, while the AVH and PsyCL datasets involved participants with more established schizophrenia spectrum disorders. This difference in illness trajectory, alongside task structure, could contribute to the observed variations in speech patterns. In the picture description task used in TOPSY, the focus on the picture provided tends to constrain speakers' responses, leading to less variability in coherence measures. In contrast, open-ended tasks, like those used to develop the AVH and PsyCL datasets, provide more variability in semantic coherence. The unstructured nature of these narratives allows for a more spontaneous expression of thought disorder, with greater fluctuations in semantic coherence that may provide a more detailed reflection of the underlying cognitive disruptions. These task-dependent and potentially illness-stage-related differences suggest that both the mode of speech elicitation and the clinical characteristics of the cohort can significantly influence the manifestation and detection of disorganized speech (Cohen et al., 2016; Parola, Lin, et al., 2023). Accordingly, interpretations of task-specific effects should be made cautiously and are intended to motivate future studies that apply multiple speech tasks within the same clinical population to better disentangle task structure from illness-related factors. While structured tasks facilitate standardized scoring and control over extraneous variables, they may inadvertently mask the intrinsic variability associated with formal thought disorder. Open-ended tasks, on the other hand, reveal the natural flow of speech, which may elicit the full spectrum of disorganization as speakers struggle to organize their thoughts in a less constrained setting. Leveraging such narrative data in future analyses could enhance the sensitivity and ecological validity of automated diagnostic tools, ultimately improving our ability to detect and characterize thought disorder in clinical populations.

## 5. Limitations and future directions

Several limitations warrant consideration. First, the heterogeneity of FTD scores (TALD, TLI, TLC) complicates direct comparisons across datasets, though our task-specific analyses mitigate this issue to some degree. Future work using harmonized rating scales across multiple speech tasks will be necessary to isolate these factors. Second, the small sample sizes in all three datasets may limit the generalizability of findings and preclude robust statistical validation of differences in model performance. In particular, while confidence intervals (CIs) are often used to quantify uncertainty and demonstrate performance improvements, the limited number of transcripts—and especially the small number of positive-class examples—results in wide and unstable CIs that make such statistical comparisons unreliable. Replication in larger datasets is needed to support formal CI-based comparisons. Third, our approach relies on ASR-generated transcripts with metadata from OpenAI's Whisper. While Whisper ASR achieved low error rates, its performance degraded with severe disorganization, highlighting the need for psychosis-specific fine-tuning. Finally, the integration strategy used—post-hoc averaging of predictions from separate pause and semantic-based models—may oversimplify the complex interplay between temporal and semantic features, suggesting that more sophisticated fusion methods might further enhance predictive performance.

While our models predict clinically rated FTD scores significantly above chance, a portion of the variance in the scores remains unexplained. This is likely because FTD is a multifaceted syndrome influenced by factors other than pause dynamics and semantic coherence (Andreasen & Grove, 1986; Roche et al., 2015). For example, other speech characteristics not explicitly modelled, such as varied syntactic, lexical, or prosodic features known to be relevant in assessing cognitive and thought disorders, could contribute (Elleuch et al., 2025a, 2025b; Voleti et al., 2019). In addition, unmeasured clinical factors, including pharmacological or psychological treatment status, may modulate speech production and symptom expression but were not consistently available across datasets. Furthermore, unmeasured individual factors, including other concurrent clinical symptoms like alogia or differing cognitive profiles, likely play a role (Berenbaum et al., 2008; Cohen et al., 2014). Finally, the inherent subjectivity in applying clinical rating scales and the structural limitations of these scales themselves can also account for some of this unexplained variance. Future research exploring these broader elements may improve predictive accuracy.

Future directions should also focus on enhancing both the temporal and semantic modeling of disorganized speech. In particular, research could benefit from the application of transformer-based models or other deep learning architectures that are capable of capturing the dynamic, real-time interplay between pause patterns and sentence-level coherence. Longitudinal studies are needed to assess the utility of these enhanced models in tracking changes in thought disorder severity over time. Furthermore, extending this work to include speakers of diverse linguistic and cultural backgrounds will be critical for ensuring that these diagnostic tools are broadly applicable. Finally, given adequately sized datasets, refining ASR systems through psychosis-specific fine-tuning could improve transcription accuracy for severely disorganized speech, thereby boosting the performance and reliability of multimodal assessments in clinical settings.

## 6. Conclusion

This study demonstrates that integrating ASR-derived pause features with sentence-level coherence metrics substantially improves the automated prediction of thought disorganization in clinical populations. Our results show that pause features, particularly simple, interpretable pause time summary statistics, are robust predictors of clinical disorganization and, when combined with semantic coherence measures, offering a scalable, multimodal framework for objective assessment. Importantly, the benefits of multimodal integration were observed consistently across datasets and coherence calculation approaches, supporting the robustness of this framework beyond any single modeling configuration. Our findings further reveal task-dependent patterns, which may also interact with the stage of illness represented in the datasets. While unstructured speech in the AVH dataset (participants with established SSD) links longer or more frequent pauses directly to disorganization, the structured picture description task used in TOPSY (participants with FEP) appears to elicit compensatory strategies that manifest as more frequent but shorter pauses. Such differences could reflect not only the cognitive demands of the task but also evolving speech characteristics or coping mechanisms at different phases of psychotic illness. Overall, this work provides a promising roadmap for refining automated task-adapted diagnostic tools for formal thought disorder, with the potential to inform earlier detection of imminent psychotic episodes to improve health outcomes in schizophrenia-spectrum disorders.

**Code availability**
Code for feature extraction and model training can be found: https://github.com/chenfeng1234567/pause_coherence.

**Acknowledgments**
This study was supported by NIH 1U01MH135901. Data acquisition for AVH was funded by National Institute of Mental Health (R01MH112641). Data acquisition for TOPSY was funded by Canadian Institutes of Health Research (FDN 154296) to LP and Academic Medical Organization of Southwest Ontario (Opportunities Grant 2016). Data acquisition for PsyCL was funded by NIH K23 MH130750 (ST) and the Brain and Behavior Research Foundation Young Investigator Grant (ST). We acknowledge the support of TalkBank (talkbank.org) and the DISCOURSE consortium (https://discourseinpsychosis.org/). L. Palaniyappan's research is supported by Monique H. Bourgeois Chair in Developmental Disorders and the Graham Boeckh Foundation. He receives a salary award from the Fonds de recherche du Québec-Santé (FRQS: 366934).

**Disclosures**
SXT owns equity and serves on the board and as a consultant for North Shore Therapeutics, received research funding and serves as a consultant for Winterlight Labs, is on the advisory board and owns equity for Psyrin, and serves as a consultant for Catholic Charities Neighborhood Services and LB Pharmaceuticals. LP reports personal fees for serving as chief editor from the Canadian Medical Association Journals, speaker honorarium from Janssen Canada and Otsuka Canada, SPMM Course Limited, UK; book royalties from Oxford University Press; investigator-initiated educational grants from Otsuka Canada outside the submitted work, in the last 5 years.

**Declaration of generative AI and AI-assisted technologies in the writing process**
During the preparation of this work the author(s) used Claude Sonnet 4 (Anthropic) to assist with manuscript preparation. The AI tool was used solely to improve readability, clarity, and adherence to journal formatting guidelines. All scientific content, data analysis, interpretation of results, and conclusions remain entirely the work of the authors. After using this tool/service, the author(s) reviewed and edited the content as needed and take(s) full responsibility for the content of the publication.

**Author contributions**
Feng Chen: Conceptualization, Data curation, Formal analysis, Methodology, Visualization, Writing – original draft, Writing – review and editing
Weizhe Xu: Conceptualization, Software, Methodology, Writing – review and editing
Changye Li: Conceptualization, Writing – review and editing
Serguei Pakhomov: Conceptualization, Supervision, Writing – review and editing
Alex Cohen: Conceptualization, Methodology, Investigation, Supervision, Writing – review and editing
Simran Bhola: Data curation, Investigation, Writing – review and editing
Sandy Yin: Data curation, Investigation, Writing – review and editing
Sunny X Tang: Resources, Data curation, Investigation, Writing – review and editing
Michael Mackinley: Data curation, Investigation, Writing – review and editing
Lena Palaniyappan: Conceptualization, Resources, Data curation, Supervision, Writing – review and editing
Dror Ben-Zeev: Data curation, Resources, Funding acquisition, Supervision, Writing – review and editing
Trevor Cohen: Conceptualization, Methodology, Supervision, Funding acquisition, Writing – original draft, Writing – review and editing

References

Abdi, H. (2010). Coefficient of variation. *Encyclopedia of research design*, *1*(5), 169-171.

Andreasen, N. C. (1986). Scale for the assessment of thought, language, and communication (TLC). *Schizophrenia bulletin*, *12*(3), 473.

Andreasen, N. C., & Grove, W. M. (1986). Thought, language, and communication in schizophrenia: diagnosis and prognosis. *Schizophrenia bulletin*, *12*(3), 348-359.

Angelopoulou, G., Kasselimis, D., Varkanitsa, M., Tsolakopoulos, D., Papageorgiou, G., Velonakis, G., Meier, E., Karavassilis, E., Pantoleon, V., & Laskaris, N. (2024). Investigating silent pauses in connected speech: integrating linguistic, neuropsychological, and neuroanatomical perspectives across narrative tasks in post-stroke aphasia. *Frontiers in Neurology*, *15*, 1347514.

Bain, M., Huh, J., Han, T., & Zisserman, A. (2023). Whisperx: Time-accurate speech transcription of long-form audio. *arXiv preprint arXiv:2303.00747*.

Bedi, G., Carrillo, F., Cecchi, G. A., Slezak, D. F., Sigman, M., Mota, N. B., Ribeiro, S., Javitt, D. C., Copelli, M., & Corcoran, C. M. (2015). Automated analysis of free speech predicts psychosis onset in high-risk youths. *npj Schizophrenia*, *1*(1), 1-7.

Ben-Zeev, D., Buck, B., Chander, A., Brian, R., Wang, W., Atkins, D., Brenner, C. J., Cohen, T., Campbell, A., & Munson, J. (2020). Mobile RDoC: using smartphones to understand the relationship between auditory verbal hallucinations and need for care. *Schizophrenia Bulletin Open*, *1*(1), sgaa060.

Berenbaum, H., Kerns, J. G., Vernon, L. L., & Gomez, J. J. (2008). Cognitive correlates of schizophrenia signs and symptoms: I. Verbal communication disturbances. *Psychiatry research*, *159*(1-2), 147-156.

Berisha, V., & Liss, J. M. (2024). Responsible development of clinical speech AI: Bridging the gap between clinical research and technology. *NPJ Digital Medicine*, *7*(1), 208.

Bird, S. (2006). NLTK: the natural language toolkit. Proceedings of the COLING/ACL 2006 Interactive Presentation Sessions,

Bird, S., Klein, E., & Loper, E. (2009). *Natural language processing with Python: analyzing text with the natural language toolkit*. " O'Reilly Media, Inc.".

Bredin, H. (2023). pyannote. audio 2.1 speaker diarization pipeline: principle, benchmark, and recipe. 24th INTERSPEECH Conference (INTERSPEECH 2023),

Christ, M., Braun, N., Neuffer, J., & Kempa-Liehr, A. W. (2018). Time series feature extraction on basis of scalable hypothesis tests (tsfresh–a python package). *Neurocomputing*, *307*, 72-77.

Ciampelli, S., Voppel, A., De Boer, J., Koops, S., & Sommer, I. (2023). Combining automatic speech recognition with semantic natural language processing in schizophrenia. *Psychiatry research*, *325*, 115252.

Cohen, A. S., Mitchell, K. R., Docherty, N. M., & Horan, W. P. (2016). Vocal expression in schizophrenia: Less than meets the ear. *Journal of Abnormal Psychology*, *125*(2), 299.

Cohen, A. S., Mitchell, K. R., & Elvevåg, B. (2014). What do we really know about blunted vocal affect and alogia? A meta-analysis of objective assessments. *Schizophrenia research*, *159*(2-3), 533-538.

Çokal, D., Zimmerer, V., Turkington, D., Ferrier, N., Varley, R., Watson, S., & Hinzen, W. (2019). Disturbing the rhythm of thought: Speech pausing patterns in schizophrenia, with and without formal thought disorder. *PloS one*, *14*(5), e0217404.

Corcoran, C. M., Carrillo, F., Fernández-Slezak, D., Bedi, G., Klim, C., Javitt, D. C., Bearden, C. E., & Cecchi, G. A. (2018). Prediction of psychosis across protocols and risk cohorts using automated language analysis. *World Psychiatry*, *17*(1), 67-75.

Corcoran, C. M., & Cecchi, G. A. (2020). Using language processing and speech analysis for the identification of psychosis and other disorders. *Biological Psychiatry: Cognitive Neuroscience and Neuroimaging*, *5*(8), 770-779.

Corcoran, C. M., Mittal, V. A., Bearden, C. E., Gur, R. E., Hitczenko, K., Bilgrami, Z., Savic, A., Cecchi, G. A., & Wolff, P. (2020). Language as a biomarker for psychosis: a natural language processing approach. *Schizophrenia research*, *226*, 158-166.

Corona Hernández, H., Corcoran, C., Achim, A. M., De Boer, J. N., Boerma, T., Brederoo, S. G., Cecchi, G. A., Ciampelli, S., Elvevåg, B., & Fusaroli, R. (2023). Natural language processing markers for psychosis and other psychiatric disorders: emerging themes and research agenda from a cross-linguistic workshop. *Schizophrenia bulletin*, *49*(Supplement_2), S86-S92.

Dalal, T. C., Liang, L., Silva, A. M., Mackinley, M., Voppel, A., & Palaniyappan, L. (2025). Speech based natural language profile before, during and after the onset of psychosis: A cluster analysis. *Acta Psychiatrica Scandinavica*, *151*(3), 332-347.

De Boer, J., Voppel, A., Brederoo, S., Schnack, H., Truong, K., Wijnen, F., & Sommer, I. (2023). Acoustic speech markers for schizophrenia-spectrum disorders: a diagnostic and symptom-recognition tool. *Psychological medicine*, *53*(4), 1302-1312.

Deerwester, S., Dumais, S. T., Furnas, G. W., Landauer, T. K., & Harshman, R. (1990). Indexing by latent semantic analysis. *Journal of the American society for information science*, *41*(6), 391-407.

Devlin, J., Chang, M.-W., Lee, K., & Toutanova, K. (2019). Bert: Pre-training of deep bidirectional transformers for language understanding. Proceedings of the 2019 conference of the North American chapter of the association for computational linguistics: human language technologies, volume 1 (long and short papers),

El Boukkouri, H., Ferret, O., Lavergne, T., & Zweigenbaum, P. (2019). Embedding strategies for specialized domains: Application to clinical entity recognition. Proceedings of the 57th Annual Meeting of the Association for Computational Linguistics: Student Research Workshop,

Elleuch, D., Chen, Y., Luo, Q., & Palaniyappan, L. (2025a). Relationship between grammar and schizophrenia: a systematic review and meta-analysis. *Communications Medicine*, *5*(1), 1-14.

Elleuch, D., Chen, Y., Luo, Q., & Palaniyappan, L. (2025b). Speaking of yourself: A meta-analysis of 80 years of research on pronoun use in schizophrenia. *Schizophrenia research*, *279*, 22-30.

Elvevåg, B., Foltz, P. W., Weinberger, D. R., & Goldberg, T. E. (2007). Quantifying incoherence in speech: an automated methodology and novel application to schizophrenia. *Schizophrenia research*, *93*(1-3), 304-316.

Foltz, P. W., Kintsch, W., & Landauer, T. K. (1998). The measurement of textual coherence with latent semantic analysis. *Discourse processes*, *25*(2-3), 285-307.

Gao, T., Yao, X., & Chen, D. (2021). Simcse: Simple contrastive learning of sentence embeddings. *arXiv preprint arXiv:2104.08821*.

Hansen, L., Rocca, R., Simonsen, A., Olsen, L., Parola, A., Bliksted, V., Ladegaard, N., Bang, D., Tylén, K., & Weed, E. (2023). Speech-and text-based classification of neuropsychiatric conditions in a multidiagnostic setting. *Nature Mental Health*, *1*(12), 971-981.

Hart, M., & Lewine, R. R. (2017). Rethinking thought disorder. In (Vol. 43, pp. 514-522): Oxford University Press US.

Hitczenko, K., Cowan, H., Mittal, V., & Goldrick, M. (2021). Automated coherence measures fail to index thought disorder in individuals at risk for psychosis. Proceedings of the seventh workshop on computational linguistics and clinical psychology: improving access,

Huang, W. R., Zhang, H., Kumar, S., Chang, S.-y., & Sainath, T. N. (2023). Semantic segmentation with bidirectional language models improves long-form asr. *arXiv preprint arXiv:2305.18419*.

Just, S., Haegert, E., Kořánová, N., Bröcker, A.-L., Nenchev, I., Funcke, J., Montag, C., & Stede, M. (2019). Coherence models in schizophrenia. Proceedings of the Sixth Workshop on Computational Linguistics and Clinical Psychology,

Just, S. A., Haegert, E., Kořánová, N., Bröcker, A.-L., Nenchev, I., Funcke, J., Heinz, A., Bermpohl, F., Stede, M., & Montag, C. (2020). Modeling incoherent discourse in non-affective psychosis. *Frontiers in Psychiatry*, *11*, 846.

Kircher, T., Krug, A., Stratmann, M., Ghazi, S., Schales, C., Frauenheim, M., Turner, L., Fährmann, P., Hornig, T., & Katzev, M. (2014). A rating scale for the assessment of objective and subjective formal Thought and Language Disorder (TALD). *Schizophrenia research*, *160*(1-3), 216-221.

Krivokapić, J., Styler, W., & Parrell, B. (2020). Pause postures: The relationship between articulation and cognitive processes during pauses. *Journal of Phonetics*, *79*, 100953.

Lesh, T. A., Niendam, T. A., Minzenberg, M. J., & Carter, C. S. (2011). Cognitive control deficits in schizophrenia: mechanisms and meaning. *Neuropsychopharmacology*, *36*(1), 316-338.

Li, C., Xu, W., Cohen, T., & Pakhomov, S. (2024). Useful blunders: Can automated speech recognition errors improve downstream dementia classification? *Journal of biomedical informatics*, *150*, 104598.

Liddle, P. F., Ngan, E. T., Caissie, S. L., Anderson, C. M., Bates, A. T., Quested, D. J., White, R., & Weg, R. (2002). Thought and Language Index: an instrument for assessing thought and language in schizophrenia. *The British Journal of Psychiatry*, *181*(4), 326-330.

Low, D. M., Bentley, K. H., & Ghosh, S. S. (2020). Automated assessment of psychiatric disorders using speech: A systematic review. *Laryngoscope investigative otolaryngology*, *5*(1), 96-116.

Matthews, N. L., Collins, K. P., Thakkar, K. N., & Park, S. (2014). Visuospatial imagery and working memory in schizophrenia. *Cognitive neuropsychiatry*, *19*(1), 17-35.

Matzinger, T., Pleyer, M., & Żywiczyński, P. (2023). Pause Length and Differences in Cognitive State Attribution in Native and Non-Native Speakers. *Languages*, *8*(1), 26.

Mikolov, T., Chen, K., Corrado, G., & Dean, J. (2013). Efficient estimation of word representations in vector space. *arXiv preprint arXiv:1301.3781*.

Nir, Y., & Tononi, G. (2010). Dreaming and the brain: from phenomenology to neurophysiology. *Trends in cognitive sciences*, *14*(2), 88-100.

Parola, A., Lin, J. M., Simonsen, A., Bliksted, V., Zhou, Y., Wang, H., Inoue, L., Koelkebeck, K., & Fusaroli, R. (2023). Speech disturbances in schizophrenia: Assessing cross-linguistic generalizability of NLP automated measures of coherence. *Schizophrenia research*, *259*, 59-70.

Parola, A., Rybner, A., Jessen, E., Mortensen, M. D., Larsen, S. N., Simonsen, A., Zhou, Y., Koelkebeck, K., Bliksted, V., & Fusaroli, R. (2023). Vocal markers of schizophrenia: assessing the generalizability of machine learning models and their clinical applicability. *European Psychiatry*, *66*(S1), S186-S186.

Parola, A., Simonsen, A., Bliksted, V., & Fusaroli, R. (2020). Voice patterns in schizophrenia: A systematic review and Bayesian meta-analysis. *Schizophrenia research*, *216*, 24-40.

Radford, A., Kim, J. W., Xu, T., Brockman, G., McLeavey, C., & Sutskever, I. (2023). Robust speech recognition via large-scale weak supervision. International conference on machine learning,

Roche, E., Creed, L., MacMahon, D., Brennan, D., & Clarke, M. (2015). The epidemiology and associated phenomenology of formal thought disorder: a systematic review. *Schizophrenia bulletin*, *41*(4), 951-962.

Sass, L., & Parnas, J. (2017). Thought disorder, subjectivity, and the self. *Schizophrenia bulletin*, *43*(3), 497-502.

Sharpe, V., Mackinley, M., Eddine, S. N., Wang, L., Palaniyappan, L., & Kuperberg, G. R. (2025). Selective insensitivity to global vs. local linguistic context in speech produced by patients with untreated psychosis and positive thought disorder. *Biological Psychiatry*.

Stanislawski, E. R., Bilgrami, Z. R., Sarac, C., Garg, S., Heisig, S., Cecchi, G. A., Agurto, C., & Corcoran, C. M. (2021). Negative symptoms and speech pauses in youths at clinical high risk for psychosis. *npj Schizophrenia*, *7*(1), 3.

Tan, E. J., Sommer, I. E., & Palaniyappan, L. (2023). Language and psychosis: Tightening the association. *Schizophrenia bulletin*, *49*(Supplement_2), S83-S85.

Tang, S. X., Hänsel, K., Cong, Y., Nikzad, A. H., Mehta, A., Cho, S., Berretta, S., Behbehani, L., Pradhan, S., & John, M. (2023). Latent factors of language disturbance and relationships to quantitative speech features. *Schizophrenia bulletin*, *49*(Supplement_2), S93-S103.

Tang, S. X., Kriz, R., Cho, S., Park, S. J., Harowitz, J., Gur, R. E., Bhati, M. T., Wolf, D. H., Sedoc, J., & Liberman, M. Y. (2021). Natural language processing methods are sensitive to sub-clinical linguistic differences in schizophrenia spectrum disorders. *npj Schizophrenia*, *7*(1), 25.

Thakore, J., Rapcan, V., & Darcy, S. (2010). Acoustic and temporal analysis of speech: a potential biomarker for schizophrenia. *Medical Engineering & Physics*, *32*(9), 1074-1079.

Van Dyken, P. C., MacKinley, M., Khan, A. R., & Palaniyappan, L. (2024). Cortical Network Disruption is Minimal in Early Stages of Psychosis. *Schizophrenia Bulletin Open*, *5*(1), sgae010.

Voleti, R., Liss, J. M., & Berisha, V. (2019). A review of automated speech and language features for assessment of cognitive and thought disorders. *IEEE journal of selected topics in signal processing*, *14*(2), 282-298.

Weizhe Xu, T. C. (2022). *Coherence*. https://github.com/LinguisticAnomalies/Coherence

Xu, W., Portanova, J., Chander, A., Ben-Zeev, D., & Cohen, T. (2021). The centroid cannot hold: comparing sequential and global estimates of coherence as indicators of formal thought disorder. AMIA Annual Symposium Proceedings,

Xu, W., Wang, W., Portanova, J., Chander, A., Campbell, A., Pakhomov, S., Ben-Zeev, D., & Cohen, T. (2022). Fully automated detection of formal thought disorder with Time-series Augmented Representations for Detection of Incoherent Speech (TARDIS). *Journal of biomedical informatics*, *126*, 103998.